\documentclass[11pt]{article}
\usepackage{amsfonts}

\usepackage[final]{acl}

\usepackage{times}
\usepackage{latexsym}
\usepackage{bbding}
\usepackage[most]{tcolorbox}
\usepackage{cuted}
\usepackage{setspace}

\usepackage[T1]{fontenc}

\usepackage[utf8]{inputenc}

\usepackage{booktabs}   
\usepackage{multirow}   
\usepackage{graphicx}   
\usepackage[table]{xcolor} 

\usepackage{microtype}
\usepackage[table]{xcolor}

\usepackage{amssymb}
\usepackage{pifont}
\usepackage{xcolor, colortbl}

\usepackage{inconsolata}

\usepackage{graphicx}

\definecolor{yesgreen}{RGB}{41, 217, 85} 
\definecolor{nored}{RGB}{214, 43, 17} 
\definecolor{mygray}{gray}{0.92}

\definecolor{acadblue}{RGB}{235, 245, 250} 
\definecolor{headerblue}{RGB}{214, 230, 245} 

\definecolor{softcyan}{RGB}{240, 250, 250} 
\definecolor{softorange}{RGB}{255, 248, 240} 
\definecolor{distinctgray}{RGB}{240, 240, 240} 

\title{OmniPhys: A Unified Multimodal Benchmark for Physics Understanding and Generation from Chinese Educational Corpora}

\author{
    \textbf{Hao Chen}\textsuperscript{1}, 
    \textbf{Yumin Lin}\textsuperscript{1}, 
    \textbf{Nadila Yushanjiang}\textsuperscript{1},
    \textbf{Xin Lin}\textsuperscript{1,$\dagger$}, 
    \textbf{Min Zhang}\textsuperscript{1,$\ddagger$,$\dagger$} 
    \\
    \textsuperscript{1}East China Normal University,
    \\
    \small \textsuperscript{$\ddagger$}Project leader \quad 
    \href{51275901038@stu.encu.edu.cn}{51275901038@stu.encu.edu.cn} \quad
    \textsuperscript{$\dagger$}Corresponding authors: \href{mailto:mzhang@cs.ecnu.edu.cn}{mzhang@cs.ecnu.edu.cn}, \href{xlin@cs.ecnu.edu.cn}{xlin@cs.ecnu.edu.cn}
}

\begin{document}
\maketitle
\begin{abstract}



Multimodal Large Language Models (MLLMs) have demonstrated strong abilities in solving diverse visual and textual reasoning tasks. 
However, their development in the physics domain is significantly hindered by the lack of a comprehensive benchmark.
To fill this gap, we introduce \textbf{OmniPhys}, a large-scale benchmark for multimodal physics understanding and reasoning, covering middle school through university-level problems from Chinese Educational Corpora. OmniPhys consists of \textbf{15,246} questions and \textbf{19,850} images, accompanied by detailed annotations that support fine-grained analysis of reasoning processes and knowledge usage.
Beyond conventional evaluation, OmniPhys is a benchmark that systematically evaluates multimodal outputs in physics domain, including models’ ability to generate structured physics diagrams, which constitute a fundamental component of authentic physics problem solving.
Extensive evaluations reveal critical gaps in the capabilities of current MLLMs, especially in complex reasoning and visual generation. To address this, we release \textbf{OmniPhys} to serve as a foundational resource for advancing multimodal intelligence in physics and scientific domains.
Codes and data are available at \url{https://github.com/ECNU-RAIL/OmniPhys-EMNLP2026}.
\end{abstract}

\section{Introduction}













The rapid evolution of Large Language Models (LLMs)~\cite{o1_system_card_2024,gpt4_technical_report_2023,gemini_2_5_2025,llama3_2024,deepseek_v3_2024} and Multimodal Large Language Models (MLLMs)~\cite{clip,flamingo,blip} has revolutionized language understanding and visual reasoning across general-purpose tasks. However, applying these models to specialized scientific domains, such as mathematical problem-solving\cite{lu2023mathvista,yang2024qwen2}, physics problem-solving\cite{PhysQA,PhysicsQA} and diagram interpretation~\cite{chartqa,plotqa}, remains challenging. \textbf{Physics}, in particular, stands out as a prototypical multimodal arena, demanding the rigorous integration of textual descriptions, visual diagrams, and symbolic logic to achieve accurate reasoning.


Physics underpins all branches of the natural sciences\cite{history_2,history_1}. 
While specific datasets have emerged to benchmark physical reasoning~\cite{PhysQA,MM-PhyQA,Ugphysics,Multi-Physics}, existing works rarely satisfy three critical criteria simultaneously: (1) \textbf{Cross-stage knowledge fusion}, spanning the full spectrum from middle school to university levels; (2) \textbf{Multimodal input comprehension}, requiring the interpretation of complex textual and visual cues; and (3) \textbf{Multimodal output generation}, assessing the model's ability to actively synthesize diagrams rather than merely selecting options. The absence of such a unified benchmark hinders a systematic evaluation of current models' capabilities.

\begin{figure*}[t]
    \centering
    \includegraphics[width=1.0\linewidth]{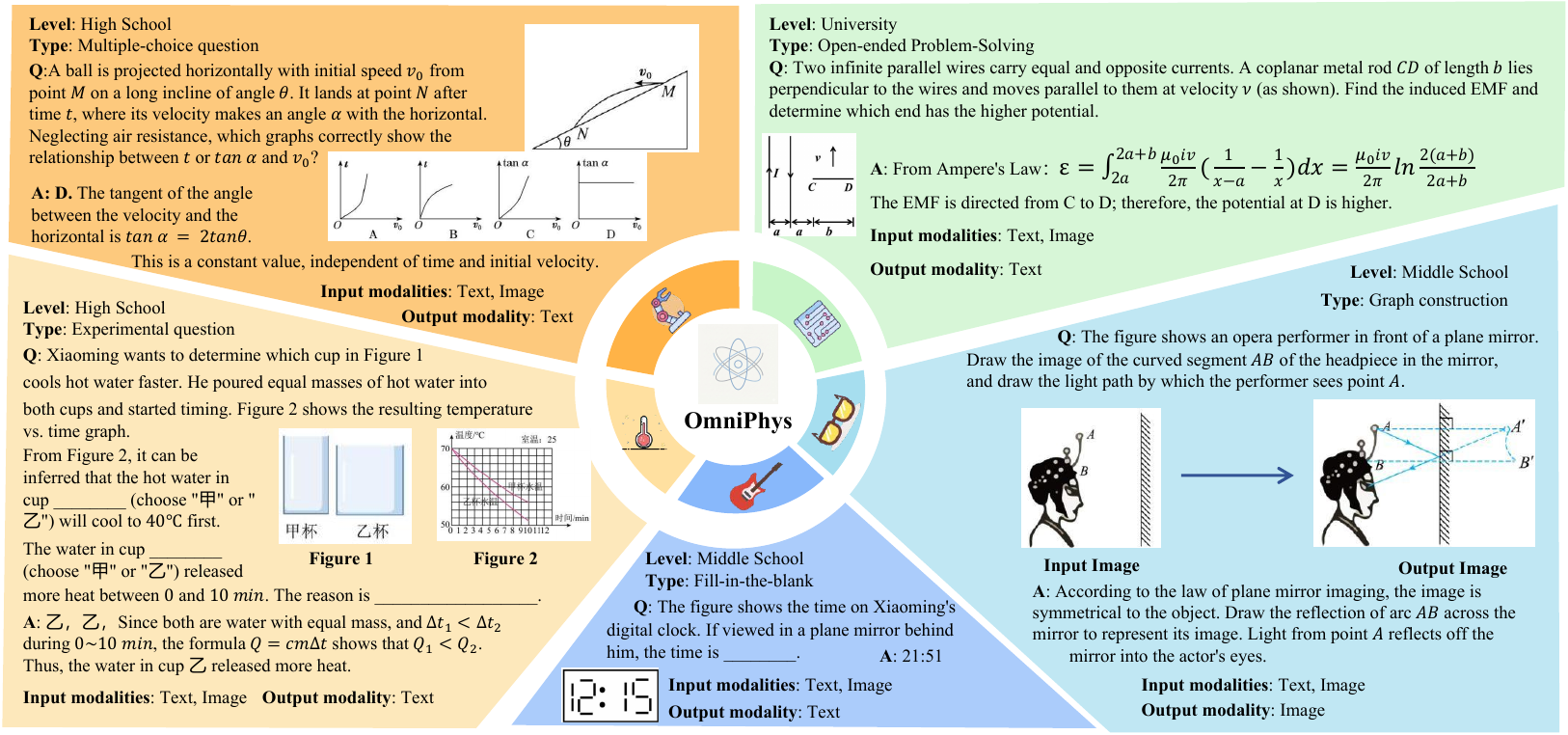}
    \caption{\textbf{Overview of the OmniPhys Dataset.} The benchmark encompasses five major physics disciplines, illustrated by representative samples: \textbf{Mechanics}, \textbf{Electromagnetism}, \textbf{Optics}, \textbf{Thermodynamics}, and \textbf{Acoustics}. 
    }
    \label{fig:overview}
\end{figure*}

To address these challenges, we present \textbf{OmniPhys}, a unified Chinese benchmark designed to assess physics mastery from secondary education to university levels. As shown in Figure \ref{fig:overview}, the dataset features a rigorous combination of textual, visual, and symbolic inputs, covering five major physical disciplines including mechanics, electromagnetism, and optics. To guarantee difficulty and pedagogical validity, all questions are meticulously curated from contemporary examination papers and authoritative textbooks, undergoing a strict multi-stage filtering process. Distinctively, OmniPhys introduces a pioneering subset for multimodal output tasks, specifically designed to assess the capabilities of MLLMs in physics diagram understanding and editing. To establish a benchmark baseline, we conducted comprehensive evaluations on OmniPhys. Our empirical results reveal that despite recent advancements, significant challenges persist for both proprietary and open-source MLLMs. Models across both categories demonstrate substantial room for improvement in handling multimodal physics reasoning across diverse educational stages.
Our main contributions are summarized as follows:

\textbf{Holistic Benchmark.} We present OmniPhys, an authentic dataset covering middle school to university physics to evaluate cross-stage reasoning transfer.

\textbf{Generative Subset.} We design a novel subset for multimodal output tasks, assessing the models' capability to synthesize and edit physics diagrams.

\textbf{Comprehensive Evaluation.} Extensive experiments with SOTA MLLMs reveal persistent challenges in conceptual mastery, positioning OmniPhys as a rigorous baseline for future research.



\section{Related Works}
\newcommand{\cmark}{\textcolor{teal}{\ding{51}}} 
\newcommand{\xmark}{\textcolor{red}{\ding{55}}} 
\definecolor{rowbg}{gray}{0.92}

\begin{table*}[htbp]
\centering
\small
\setlength{\tabcolsep}{3.5pt} 
\resizebox{\linewidth}{!}{ 
\begin{tabular}{l c c cccc cc c}
\toprule
\multirow{2}{*}{\textbf{Dataset}} & \multirow{2}{*}{\textbf{Size (Phys.)}} & \multirow{2}{*}{\textbf{Lang.}} & \multicolumn{4}{c}{\textbf{Education Stage}} & \multicolumn{2}{c}{\textbf{Image Modality}} & \multirow{2}{*}{\textbf{Overlap}} \\ 
\cmidrule(lr){4-7} \cmidrule(lr){8-9}
 & & & \textbf{PS} & \textbf{MS} & \textbf{HS} & \textbf{Uni} & \textbf{Input} & \textbf{Output} & \\
\midrule

\multicolumn{10}{l}{\textit{Text-Only Benchmarks}} \\
PhysQA~\cite{PhysQA}       & 1,008  & EN    & \xmark & \cmark & \xmark & \xmark & \xmark & \xmark & 0.00\% \\
C-Eval~\cite{ceval}        & 601    & ZH    & \xmark & \cmark & \cmark & \cmark & \xmark & \xmark & 0.00\% \\
MMLU~\cite{mmlu}           & 548    & EN    & \xmark & \xmark & \cmark & \cmark & \xmark & \xmark & 0.00\% \\
GAOKAO~\cite{GAGOKAO}      & 111    & ZH    & \xmark & \xmark & \cmark & \xmark & \xmark & \xmark & 0.00\% \\
AGIEval~\cite{Agieval}     & 200    & EN    & \xmark & \xmark & \cmark & \xmark & \xmark & \xmark & 0.00\% \\
UGPhysics~\cite{Ugphysics} & 11,040 & ZH/EN & \xmark & \xmark & \xmark & \cmark & \xmark & \xmark & 0.00\% \\
PHYSICS~\cite{PHYSICS}     & 16,568 & ZH/EN & \xmark & \xmark & \cmark & \cmark & \xmark & \xmark & 0.00\% \\

\midrule

\multicolumn{10}{l}{\textit{Multimodal Input Benchmarks}} \\
TheoremQA~\cite{Theoremqa}     & 131   & EN    & \xmark & \xmark & \xmark & \cmark & \cmark & \xmark & 0.00\% \\
Multi-Physics~\cite{Multi-Physics} & 1,412 & ZH    & \xmark & \xmark & \cmark & \xmark & \cmark & \xmark & 0.00\% \\
OlympiadBench~\cite{Olympiadbench} & 2,334 & ZH/EN & \xmark & \xmark & \cmark & \cmark & \cmark & \xmark & 0.00\% \\
MM-PhyQA~\cite{MM-PhyQA}       & 4,500 & EN    & \xmark & \xmark & \cmark & \xmark & \cmark & \xmark & 0.00\% \\
PhysicsArena~\cite{PhysicsArena} & 5,103 & EN    & \xmark & \xmark & \cmark & \xmark & \cmark & \xmark & 0.00\% \\
MM-Eureka~\cite{Mm-eureka}     & 500   & EN    & \xmark & \xmark & \cmark & \xmark & \cmark & \xmark & 0.00\% \\
PhysReason~\cite{zhang2025physreason}       & 1,200 & EN    & \xmark & \xmark & \xmark & \cmark & \cmark & \xmark & 0.00\% \\
See-Phys~\cite{xiang2025seephys}       & 2,000 & EN    & \cmark & \cmark & \cmark & \cmark & \cmark & \xmark & 0.00\% \\
K12Vista~\cite{K12Vista}       & 6,600 & EN    & \cmark & \cmark & \cmark & \xmark & \cmark & \xmark & 0.02\% \\
MDK12-Bench~\cite{mdk12}       & 8,542 & EN    & \cmark & \cmark & \cmark & \xmark & \cmark & \xmark & 0.04\% \\

\midrule

\rowcolor{gray!15}
\textbf{OmniPhys (Ours)} & \textbf{15,246} & \textbf{ZH} & \textbf{\cmark} & \textbf{\cmark} & \textbf{\cmark} & \textbf{\cmark} & \textbf{\cmark} & \textbf{\cmark} & \textbf{-} \\

\bottomrule
\end{tabular}
}
\caption{\textbf{Comparison with Existing Physics Datasets.} \textbf{OmniPhys} distinguishes itself by supporting \textit{multimodal outputs} and covering the full spectrum of educational stages. (Lang.: Language, PS: Primary School, MS: Middle School, HS: High School, Uni: University) \textit{Overlap} (rightmost column) indicates the percentage of samples in OmniPhys that overlap with existing benchmarks, computed via string similarity and perceptual hashing. Our dataset maintains near-zero overlap while uniquely supporting multimodal outputs across all stages.}
\label{tab:comparison}
\end{table*}

\subsection{Multimodal Large Language Models}

Recent MLLMs, such as LLaVA~\cite{llava} and BLIP-2~\cite{blip-2}, leverage instruction tuning to adapt LLM reasoning to multimodal contexts. While frontier models like GPT-4V~\cite{gpt-4v} and Gemini~\cite{gemini_2_5_2025} now support complex interleaved inputs and outputs, they remains susceptible to hallucinations~\cite{hallucination_1} and often falter in rigorous logical deduction or numerical calculation~\cite{mllm_reason_1,jia2026diacdm}. These persistent vulnerabilities underscore the urgent necessity for challenging benchmarks to probe the upper bounds of deep multimodal reasoning.

\subsection{Benchmark for Physics Reasoning}


Table \ref{tab:comparison} summarizes representative physical reasoning datasets. Early research primarily focused on text-only modalities \cite{PhysQA,PhysicsQA,Ugphysics,PHYSICS,qu2026tpop} or general K-12 benchmarks with limited physics subsets \cite{mmlu,cmmlu,Agieval,ceval,GAGOKAO}. While recent works have introduced image inputs \cite{Olympiadbench,Olympicarena,K12Vista,mdk12,codolzhang2025}, they still suffer from incomplete educational coverage and a lack of multimodal outputs \cite{bench-v}. 
To bridge these gaps, we introduce \textbf{OmniPhys}, a unified benchmark meticulously curated from authentic Chinese educational corpora. Unlike existing resources that frequently overlap in web-crawled content, OmniPhys maintains a near-zero overlap (<0.05\%) with current benchmarks, ensuring a non-contaminated and independent evaluation. By spanning the full spectrum from K-12 to university levels and uniquely supporting multimodal outputs, OmniPhys provides a rigorous testbed for both physics understanding and diagrammatic generation.
\section{The OmniPhys Benchmark}


We introduce \textbf{OmniPhys}, a comprehensive multimodal physics benchmark spanning junior high to university curricula. It challenges MLLMs to synergize diagrams, formulas, and text for rigorous, visually grounded reasoning.

    

\begin{table}[t]
    \centering
    \small
    \begin{tabular*}{\linewidth}{l@{\extracolsep{\fill}}r}
        \toprule
        \textbf{Category} & \textbf{Count} \\
        \midrule
        \multicolumn{2}{l}{\textit{Physics Domains}} \\
        \hspace{1em} Mechanics & 7,240 \\
        \hspace{1em} Electromagnetism & 5,325 \\
        \hspace{1em} Optics & 1,120 \\
        \hspace{1em} Thermodynamics & 856 \\
        \hspace{1em} Acoustics & 705 \\
        \midrule
        \multicolumn{2}{l}{\textit{Educational Stages}} \\
        \hspace{1em} Junior High School & 4,524 \\
        \hspace{1em} Senior High School & 9,217 \\
        \hspace{1em} University (Advanced) & 1,505 \\ 
        \midrule
        \multicolumn{2}{l}{\textit{Task Types}} \\
        \hspace{1em} Objective (e.g., MCQ) & 10,227 \\
        \hspace{1em} Open-Ended (e.g., Calculation) & 4,258 \\
        \hspace{1em} Multimodal Generation & 761 \\ 
        \midrule
        \textbf{Total Questions} & \textbf{15,246} \\ 
        \textbf{Total Images} & \textbf{19,850} \\ 
        \bottomrule
    \end{tabular*}
    \caption{\textbf{Statistics of OmniPhys.} The benchmark maintains a high-quality distribution across three educational stages. In response to the growing need for generative AI, we specifically curated a subset for \textbf{Multimodal Generation}, featuring expert-annotated diagrams.}
    \label{tab:data_stats}
\end{table}





\subsection{Data Sources and Distribution}




OmniPhys is curated from authoritative Chinese pedagogical resources and authentic examination papers, spanning middle school to university curricula. 
The source materials are collected from publicly accessible educational platforms (e.g., Zxxk.com) and open examination archives, and used under fair-use principles for non-commercial academic research. To respect intellectual property, the released benchmark includes only structured annotations (text, reasoning chains, and re-rendered diagrams), rather than verbatim source PDFs.
We utilize a multi-stage pipeline, integrating MinerU~\cite{mineru1} and DeepSeek-OCR~\cite{deepseekocr}to transform raw PDFs into a structured \texttt{JSONL} format. Each entry features comprehensive annotations, including problem texts, associated diagrams, and step-by-step reasoning chains. Representative samples of the source materials are provided in Appendix \ref{sec:appendix_samples}.

Table \ref{tab:data_stats} summarizes the composition of OmniPhys, which is strategically architected to mirror real-world physics curricula. Centered on a "difficulty pyramid," the benchmark scales from foundational K-12 concepts to advanced university-level challenges. This hierarchical structure is designed to stress-test the upper bounds of model reasoning in expert-level scenarios, moving beyond broad but shallow evaluations.

\subsection{Data Preprocessing and Quality Filtering}

To ensure the correctness, clarity, and reliability of \textbf{OmniPhys}, we implement a multi-stage data preprocessing and filtering pipeline.

\textbf{Completeness Screening.} 
Each sample is strictly required to adhere to a valid \texttt{JSON} schema and contain all mandatory fields, including the problem statement, reference answer, and detailed solution. Furthermore, we prune samples exhibiting insufficient textual length, effectively filtering out incomplete or low-quality content.

\textbf{Semantic Deduplication.} We employ an embedding-based deduplication strategy using the \texttt{bge-small-zh-v1.5} encoder~\cite{bge_embedding}. We compute dense vector representations for all problem texts and identify duplicates based on a cosine similarity threshold of 0.95. 
This rigorous process eliminated 12.1\% of redundant instances to yield the final valid dataset.

\textbf{Visual Dependency Filtering.} To ensure OmniPhys targets genuine multimodal reasoning, we categorize instances into three levels. \textbf{Level 1 (Text-Solvable)} problems contain only decorative images where all necessary information is fully specified in the text. \textbf{Level 2 (Text-Descriptive)} problems include images that convey information, but the textual description completely duplicates the visual content. \textbf{Level 3 (Image-Essential)} problems require direct interpretation of the image, as critical quantities or relationships are only available in the diagram.
To ensure robust classification, we employ a diverse judge panel consisting of DeepSeek-V3\cite{deepseek_v3_2024}, Qwen2.5\cite{qwen3technicalreport}, and GPT-3.5\cite{gpt3}. Only samples classified as \textbf{Level 3} by all three judges are admitted into the final dataset. 
The detailed prompting strategy is provided in Appendix~\ref{app:visual_dependency_prompt}.
To validate this text-based inference strategy, we conducted 
human-machine alignment on a random sample of 300 instances in Appendix \ref{sec:appendix_samples}, 
achieving 89.3\% agreement and Cohen's $\kappa$ = 0.82 with expert judgment. 
This confirms that visual dependency can be reliably inferred from textual cues alone.


\textbf{Difficulty Screening.} We implement a model-based difficulty screening. We employ two representative lightweight MLLMs: Qwen2.5-VL-3B\cite{qwen2.5-VL} and MiniCPM-V-2.6\cite{minicpm}. We adopt an adversarial filtering criterion: any problem correctly solved by both models is deemed insufficiently challenging and is subsequently discarded. 
The process is parallelized using the \texttt{vLLM} framework on 2 NVIDIA RTX 4090 GPUs. This step refines the dataset distribution, removing easy problems to better reflect meaningful cross-stage physics mastery.

\textbf{Data Leakage Prevention.}  We specifically selecting materials from frontline educators and physically digitized examinations that are effectively insulated from public indexing. To empirically verify this isolation, we adopt a search-based filtering protocol. We eliminate samples where \texttt{GPT-4}~\cite{gpt4_technical_report_2023} retrieves the exact solution via web browsing or exhibits inconsistent responses when the search function is toggled. Finally, manual verification is conducted on the remaining subset to guarantee a valid zero-shot evaluation. 


\textbf{Human Validation.}
Expert cross-validation on randomized subsets (Appendix \ref{sec:alignment}) confirms high consistency between our automated pipeline and human judgment. Meanwhile, all collected materials were manually screened to exclude personally identifying information and sensitive content.


\definecolor{colorgray}{gray}{0.92} 
\definecolor{emphgray}{gray}{0.85}  
\definecolor{mydarkblue}{rgb}{0.0, 0.0, 205.00} 

\begin{table*}[ht]
\centering
\small
\setlength{\tabcolsep}{2.0pt} 

\resizebox{\textwidth}{!}{
\begin{tabular}{l|ccc|ccc|ccc|ccc|ccc|ccc||cc}
\toprule
\multirow{2}{*}{\textbf{Model}} & 
\multicolumn{3}{c|}{\textbf{Acoustics}} & 
\multicolumn{3}{c|}{\textbf{Optics}} & 
\multicolumn{3}{c|}{\textbf{Mechanics}} & 
\multicolumn{3}{c|}{\textbf{Thermodynamics}} & 
\multicolumn{3}{c|}{\textbf{Electricity}} & 
\multicolumn{5}{c}{\textbf{Overall}} \\ 
 & $S_1$ & $S_2$ & $S_3$ & $S_1$ & $S_2$ & $S_3$ & $S_1$ & $S_2$ & $S_3$ & $S_1$ & $S_2$ & $S_3$ & $S_1$ & $S_2$ & $S_3$ & \cellcolor{colorgray}$S_1$ & \cellcolor{colorgray}$S_2$ & \cellcolor{colorgray}$S_3$ & \cellcolor{emphgray}$P_{obj}$ & \cellcolor{emphgray}$P_{open}$ \\
\midrule
\multicolumn{21}{c}{\textit{Proprietary / Closed-source MLLMs}} \\
\midrule
GPT-5.2             & 67.48 & 81.65 & 67.78 & 65.15 & 79.73 & 64.42 & 66.33 & 79.67 & 72.30 & 69.89 & 82.13 & 73.84 & 60.15 & 77.05 & 66.34 & \cellcolor{colorgray}64.15 & \cellcolor{colorgray}78.85 & \cellcolor{colorgray}69.42 & \cellcolor{emphgray}39.36 & \cellcolor{emphgray}29.45 \\

GPT-5.1             & 58.68 & 72.24 & 55.40 & 52.77 & 66.09 & 49.32 & 52.30 & 67.91 & 54.47 & 59.87 & 73.50 & 52.04 & 45.48 & 64.07 & 50.95 & \cellcolor{colorgray}50.25 & \cellcolor{colorgray}66.68 & \cellcolor{colorgray}52.59 & \cellcolor{emphgray}23.53 & \cellcolor{emphgray}12.87 \\

GPT-4o              & 46.21 & 64.07 & 47.36 & 41.15 & 55.99 & 35.03 & 39.41 & 54.01 & 40.33 & 49.02 & 60.46 & 46.76 & 36.63 & 52.24 & 38.28 & \cellcolor{colorgray}39.04 & \cellcolor{colorgray}53.93 & \cellcolor{colorgray}39.50 & \cellcolor{emphgray}12.28 & \cellcolor{emphgray}4.61 \\

o4-mini             & 62.39 & 68.86 & 58.46 & 59.19 & 67.79 & 49.08 & 64.85 & 72.01 & 60.01 & 70.76 & 77.39 & 63.86 & 59.55 & 67.78 & 52.63 & \cellcolor{colorgray}62.75 & \cellcolor{colorgray}70.36 & \cellcolor{colorgray}56.48 & \cellcolor{emphgray}39.34 & \cellcolor{emphgray}21.50 \\

Claude-4.5-Sonnet   & 59.56 & 72.98 & 66.46 & 56.50 & 72.72 & 57.27 & 54.24 & 71.73 & 62.99 & 56.07 & 69.90 & 64.02 & 45.44 & 65.11 & 59.14 & \cellcolor{colorgray}51.31 & \cellcolor{colorgray}69.30 & \cellcolor{colorgray}61.11 & \cellcolor{emphgray}28.42 & \cellcolor{emphgray}24.42 \\

Gemini-3-Pro        & \underline{77.89} & 80.00 & \underline{78.92} & \textcolor{mydarkblue}{\textbf{79.83}} & \textcolor{mydarkblue}{\textbf{82.01}} & \textcolor{mydarkblue}{\textbf{69.81}} & \textcolor{mydarkblue}{\textbf{83.63}} & \underline{85.30} & \textcolor{mydarkblue}{\textbf{80.63}} & \textcolor{mydarkblue}{\textbf{86.51}} & \textcolor{mydarkblue}{\textbf{88.43}} & \textcolor{mydarkblue}{\textbf{81.47}} & \textcolor{mydarkblue}{\textbf{78.86}} & 80.84 & \textcolor{mydarkblue}{\textbf{73.71}} & \cellcolor{colorgray}\textcolor{mydarkblue}{\textbf{81.66}} & \cellcolor{colorgray}\underline{83.49} & \cellcolor{colorgray}\textcolor{mydarkblue}{\textbf{77.15}} & \cellcolor{emphgray}\textcolor{mydarkblue}{\textbf{69.82}} & \cellcolor{emphgray}\textcolor{mydarkblue}{\textbf{49.30}} \\

Gemini-2.5-Flash    & 67.57 & 81.86 & 77.00 & 68.24 & \underline{81.81} & \underline{64.54} & 69.50 & 84.33 & \underline{74.49} & 78.07 & \underline{87.42} & \underline{75.19} & 64.56 & \underline{80.98} & \underline{70.33} & \cellcolor{colorgray}67.93 & \cellcolor{colorgray}83.01 & \cellcolor{colorgray}\underline{72.20} & \cellcolor{emphgray}46.31 & \cellcolor{emphgray}\underline{35.89} \\

Grok-4              & 48.30 & 65.13 & 59.33 & 50.45 & 67.70 & 50.55 & 47.91 & 66.14 & 51.76 & 56.13 & 70.63 & 54.89 & 42.64 & 62.82 & 48.60 & \cellcolor{colorgray}46.53 & \cellcolor{colorgray}65.21 & \cellcolor{colorgray}50.60 & \cellcolor{emphgray}18.96 & \cellcolor{emphgray}13.80 \\

Doubao-Seed-1.6     & \textcolor{mydarkblue}{\textbf{79.80}} & \textbf{84.60} & \textcolor{mydarkblue}{\textbf{81.66}} & \underline{74.75} & 81.04 & 59.34 & \underline{82.64} & \textcolor{mydarkblue}{\textbf{86.51}} & 72.03 & \underline{82.35} & 87.21 & 75.05 & \underline{75.21} & \textcolor{mydarkblue}{\textbf{81.75}} & 68.95 & \cellcolor{colorgray}\underline{79.35} & \cellcolor{colorgray}\textcolor{mydarkblue}{\textbf{84.41}} & \cellcolor{colorgray}70.16 & \cellcolor{emphgray}\underline{61.42} & \cellcolor{emphgray}32.80 \\

GLM-4.6v            & 70.85 & 79.05 & 73.42 & 69.24 & 78.67 & 62.24 & 75.05 & 82.59 & 70.98 & 76.30 & 83.15 & 71.81 & 68.16 & 77.48 & 66.10 & \cellcolor{colorgray}72.13 & \cellcolor{colorgray}80.44 & \cellcolor{colorgray}68.49 & \cellcolor{emphgray}52.12 & \cellcolor{emphgray}33.14 \\

Qwen3-VL-Plus       & 68.71 & \underline{83.98} & 74.66 & 67.71 & 80.10 & 52.86 & 70.98 & 83.01 & 68.32 & 75.71 & 84.05 & 67.86 & 65.55 & 79.02 & 65.90 & \cellcolor{colorgray}68.96 & \cellcolor{colorgray}81.42 & \cellcolor{colorgray}66.34 & \cellcolor{emphgray}45.58 & \cellcolor{emphgray}29.60 \\


\midrule
\multicolumn{21}{c}{\textit{Open-source MLLMs}} \\
\midrule

Qwen2.5-VL-3B       & 29.97 & 28.59 & 16.83 & 23.65 & 21.62 & 5.84 & 24.06 & 18.63 & 10.04 & 29.71 & 26.56 & 12.71 & 21.57 & 16.16 & 9.92 & \cellcolor{colorgray}23.45 & \cellcolor{colorgray}18.42 & \cellcolor{colorgray}9.88 & \cellcolor{emphgray}0.96 & \cellcolor{emphgray} 0.11 \\

Qwen2.5-VL-7B       & 46.50 & 40.30 & 26.42 & 40.44 & 41.56 & 14.96 & 39.72 & 39.00 & 20.94 & 44.44 & 46.55 & 22.24 & 36.92 & 37.29 & 19.45 & \cellcolor{colorgray}39.05 & \cellcolor{colorgray}38.89 & \cellcolor{colorgray}20.03 & \cellcolor{emphgray}6.74 & \cellcolor{emphgray} 1.21 \\

Qwen2.5-VL-32B      & 56.32 & 68.27 & 60.37 & 53.13 & 65.41 & 37.29 & 56.54 & 68.18 & 56.13 & 63.63 & 72.71 & 53.05 & 48.79 & 62.34 & 49.31 & \cellcolor{colorgray}53.79 & \cellcolor{colorgray}66.05 & \cellcolor{colorgray}51.99 & \cellcolor{emphgray}25.42 & \cellcolor{emphgray} 14.08 \\

Qwen2.5-VL-72B      & 60.98 & 69.80 & 60.42 & 58.11 & 66.42 & 35.20 & 61.16 & 67.91 & 48.41 & 70.04 & 74.20 & 51.03 & 55.30 & 63.98 & 45.18 & \cellcolor{colorgray}59.19 & \cellcolor{colorgray}66.67 & \cellcolor{colorgray}46.41 & \cellcolor{emphgray}26.40 & \cellcolor{emphgray} 9.61 \\

Qwen3-VL-2B         & 39.36 & 52.22 & 29.08 & 37.55 & 50.22 & 20.84 & 31.70 & 48.45 & 26.78 & 41.00 & 55.26 & 25.26 & 29.97 & 46.60 & 25.59 & \cellcolor{colorgray}32.00 & \cellcolor{colorgray}48.25 & \cellcolor{colorgray}25.84 & \cellcolor{emphgray}6.82 & \cellcolor{emphgray}2.61 \\

Qwen3-VL-8B         & 54.88 & 69.63 & 46.58 & 52.79 & 67.33 & 31.26 & 51.83 & 67.92 & 43.40 & 60.34 & 71.91 & 46.41 & 46.77 & 65.25 & 42.39 & \cellcolor{colorgray}50.43 & \cellcolor{colorgray}67.08 & \cellcolor{colorgray}42.35 & \cellcolor{emphgray}22.90 & \cellcolor{emphgray}7.93 \\

Qwen3-VL-32B        & \underline{66.03} & \underline{77.92} & \textbf{70.00} & \underline{63.97} & \underline{75.81} & \underline{48.42} & \underline{64.68} & \underline{78.03} & \underline{64.20} & \underline{72.70} & \underline{80.76} & \underline{64.25} & \underline{57.17} & \underline{73.60} & \underline{60.21} & \cellcolor{colorgray}\underline{62.25} & \cellcolor{colorgray}\underline{76.36} & \cellcolor{colorgray}\underline{61.57} & \cellcolor{emphgray}\underline{36.08} & \cellcolor{emphgray}\underline{22.57} \\

Qwen3-VL-235B-a22b  & \textbf{74.57} & \textcolor{mydarkblue}{\textbf{84.87}} & \underline{69.96} & \textbf{68.45} & \textbf{79.64} & \textbf{54.37} & \textbf{70.72} & \textbf{83.19} & \textbf{69.13} & \textbf{75.02} & \textbf{85.03} & \textbf{69.99} & \textbf{66.12} & \textbf{79.53} & \textbf{66.30} & \cellcolor{colorgray}\textbf{69.15} & \cellcolor{colorgray}\textbf{81.72} & \cellcolor{colorgray}\textbf{67.05} & \cellcolor{emphgray}\textbf{47.16} & \cellcolor{emphgray}\textbf{29.01} \\

Kimi-VL-A3B         & 41.39 & 45.54 & 26.75 & 37.20 & 43.55 & 16.03 & 35.28 & 39.47 & 21.88 & 43.19 & 48.39 & 23.41 & 32.13 & 38.75 & 21.09 & \cellcolor{colorgray}34.70 & \cellcolor{colorgray}39.97 & \cellcolor{colorgray}21.28 & \cellcolor{emphgray}5.61 & \cellcolor{emphgray} 0.79 \\

InternVL3.5-2B      & 30.82 & 42.24 & 28.61 & 31.79 & 41.00 & 12.92 & 25.98 & 35.47 & 18.73 & 34.12 & 43.84 & 18.29 & 23.75 & 34.55 & 17.88 & \cellcolor{colorgray}25.99 & \cellcolor{colorgray}35.98 & \cellcolor{colorgray}18.04 & \cellcolor{emphgray}3.06 & \cellcolor{emphgray}0.53 \\

InternVL3.5-14B     & 50.40 & 55.46 & 46.34 & 47.59 & 54.47 & 22.79 & 40.95 & 52.77 & 32.88 & 52.51 & 59.89 & 36.11 & 36.23 & 49.27 & 30.20 & \cellcolor{colorgray}40.31 & \cellcolor{colorgray}51.96 & \cellcolor{colorgray}31.36 & \cellcolor{emphgray}12.75 & \cellcolor{emphgray}3.68 \\

InternVL3.5-38B     & 51.02 & 63.30 & 39.46 & 47.61 & 60.75 & 28.65 & 41.91 & 57.22 & 38.09 & 57.15 & 65.45 & 39.23 & 37.29 & 53.33 & 35.67 & \cellcolor{colorgray}41.41 & \cellcolor{colorgray}56.49 & \cellcolor{colorgray}36.54 & \cellcolor{emphgray}13.32 & \cellcolor{emphgray}4.10 \\

Deepseek-VL-7B      & 24.09 & 13.66 & 11.92 & 23.47 & 15.23 & 7.56 & 22.25 & 12.35 & 6.25 & 24.54 & 17.35 & 9.72 & 22.89 & 13.48 & 7.92 & \cellcolor{colorgray}22.69 & \cellcolor{colorgray}13.19 & \cellcolor{colorgray}7.21 & \cellcolor{emphgray}1.06 & \cellcolor{emphgray} 0.14 \\

LLaVA-OneVision-1.5-8B & 38.26 & 48.45 & 30.38 & 38.50 & 44.77 & 21.81 & 36.23 & 39.14 & 27.42 & 41.92 & 43.40 & 29.94 & 33.43 & 40.77 & 26.01 & \cellcolor{colorgray}35.65 & \cellcolor{colorgray}40.45 & \cellcolor{colorgray}26.61 & \cellcolor{emphgray}7.26 & \cellcolor{emphgray} 2.53 \\

Phi-4-Multimodal    & 17.24 & 18.50 & 12.91 & 19.76 & 16.83 & 5.31 & 18.71 & 14.24 & 5.23 & 21.43 & 20.67 & 5.72 & 20.23 & 15.18 & 6.20 & \cellcolor{colorgray}19.44 & \cellcolor{colorgray}15.10 & \cellcolor{colorgray}5.70 & \cellcolor{emphgray}0.53 & \cellcolor{emphgray} 0.14 \\

\bottomrule
\end{tabular}
}
\caption{\textbf{Main Results on OmniPhys.} Metrics: $S_1$ (Answer Accuracy), $S_2$/$S_3$ (Process Quality for Objective/Open-ended tasks), and $P_{obj}$, $P_{open}$ (Strict Mastery Rates). \textbf{Formatting}: In each section (Closed-source MLLMs/Open-source MLLMs), the \textbf{best} score is bolded and the \underline{second best} is underlined. The \textbf{global best} across all models is highlighted in \textcolor{mydarkblue}{\textbf{blue}}.}
\label{tab:main_results}
\end{table*}

\section{Experiments}

\subsection{Experimental Setup}

We evaluate a diverse suite of state-of-the-art MLLMs on \textbf{OmniPhys}, spanning both proprietary and open-source categories. The models included in the evaluation are detailed below.

\textit{Proprietary Models.} We prioritize recently released models to benchmark the upper bounds of current capabilities. Our evaluation covers a broad spectrum of high-performance systems, including GPT-5.2\cite{openai2025gpt52}, GPT-5.1, GPT-4o, and o4-mini, alongside competitive counterparts such as Gemini-3-pro\cite{gemini-3}, Gemini-2.5-Flash\cite{gemini_2_5_2025}, Claude-4.5-sonnet\cite{claude4.5},  Grok-4\cite{grok}, Doubao-Seed-1.6\cite{seed-1.6}, GLM-4.6v\cite{glm}, and Qwen3-VL-Plus\cite{qwen3vl}. All proprietary models are accessed via their official APIs.

\textit{Open-source Models.} 
We assess the full spectrum of the Qwen2.5-VL\cite{qwen3vl} series (3B, 7B, 32B, 72B) and its successor Qwen3-VL (2B, 8B, 32B, 235B), alongside InternVL-3.5\cite{internvl3.5} at 2B, 14B, and 38B scales. To further ensure architectural diversity, we incorporate representative models such as Kimi-VL-A3B\cite{kimiteam2025kimivltechnicalreport}, DeepSeek-VL-7B\cite{lu2024deepseekvl}, LLaVA-OneVision-1.5-8B\cite{an2025llavaonevision15fullyopenframework}, and Phi-4-Multimodal\cite{abdin2024phi4technicalreport}. The Qwen3-VL-235B-a22b is accessed via its official API, while all other open-source models are evaluated locally using 8 NVIDIA RTX 4090 and 2 NVIDIA RTX PRO 6000 GPUs, under identical inference settings.

\subsection{Dual-Track Reasoning Evaluation}
To achieve a granular assessment of multimodal reasoning capabilities, we propose the \textbf{Dual-Track Reasoning Evaluation (DTRE)} framework. We categorize problems into two distinct streams: Objective Tasks with deterministic outputs (e.g., multiple-choice, fill-in-the-blank) and Open-Ended Tasks requiring step-by-step derivation (e.g., calculation, experimental design).

To quantify model performance, we formalize two complementary metrics: \textbf{Result Score} ($S_{res}$) and \textbf{Process Score} ($S_{proc}$).

\begin{figure*}[h]
    \centering
    \begin{minipage}[b]{0.48\linewidth}
        \centering
        \includegraphics[width=\linewidth]{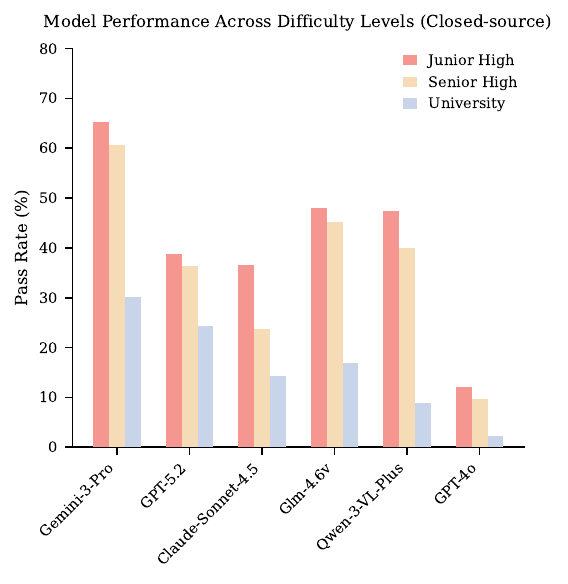}
    \end{minipage}
    \hfill 
    \begin{minipage}[b]{0.48\linewidth}
        \centering
        \includegraphics[width=\linewidth]{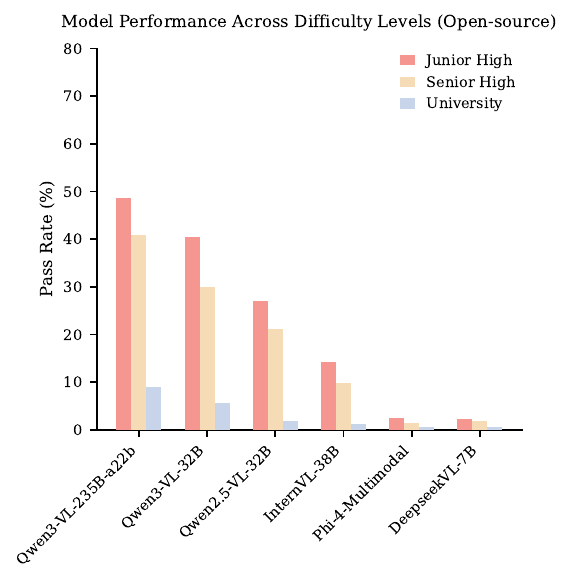}
    \end{minipage}
    
    \caption{\textbf{Performance Degradation Across Educational Stages.} We report the pass rates (\%) of 12 representative MLLMs across Junior High, Senior High, and University levels. Both closed-source (Left) and open-source (Right) models exhibit a consistent downward trend, validating the hierarchical difficulty design of OmniPhys.}
    \label{fig:model_performance}
\end{figure*}

\paragraph{Result Score ($S_{res}$)}
Designed for \textbf{Objective Tasks}, this metric quantifies the accuracy of the final answer. We denote the ground truth set as $\mathcal{A}_{gt}$ and the predicted answer set as $\mathcal{A}_{pred}$. To ensure robustness against formatting variations, we apply a standard normalization function $\phi(\cdot)$.

For single-choice questions and single-slot fill-in-the-blank tasks, the scoring is binary:
\begin{equation}
    S_{res} = \mathbb{I}(\phi(\mathcal{A}_{pred}) = \phi(\mathcal{A}_{gt}))
\end{equation}
where $\mathbb{I}(\cdot)$ is the indicator function.

For multi-select questions and multi-slot tasks, we adopt a \textbf{strict partial credit mechanism} to penalize random guessing. A score is awarded if and only if the predicted set is a subset of the ground truth (i.e., no incorrect options are selected). The score is calculated as:
\begin{equation}
    S_{res} = \begin{cases} 
    \frac{|\phi(\mathcal{A}_{pred})|}{|\phi(\mathcal{A}_{gt})|} & \text{if } \phi(\mathcal{A}_{pred}) \subseteq \phi(\mathcal{A}_{gt}) \\
    0 & \text{otherwise}
    \end{cases}
\end{equation}

\paragraph{Process Score ($S_{proc}$)}
To quantitatively assess the quality of the Chain-of-Thought (CoT) in Open-Ended Tasks, we define a metric based on the completeness of the logical derivation. Let the ground truth reasoning path be decomposed into a set of $M$ \textbf{key reasoning steps}, denoted as $\mathcal{K} = \{k_1, k_2, ..., k_M\}$. We verify whether each key step $k_i$ is explicitly or implicitly present and correctly applied in the model's reasoning path $\mathcal{R}_{pred}$.
The process score is calculated as the recall rate of these key steps:
\begin{equation}
    S_{proc} = \frac{1}{M} \sum_{i=1}^{M} \delta(k_i, \mathcal{R}_{pred})
\end{equation}
where $\delta(k_i, \mathcal{R}_{pred}) \in \{0, 1\}$ indicates whether the $i$-th key step is successfully recovered in the model's generation. 
we denote the Result Score on objective tasks as $S_1$, the Process Score (Sproc) on objective tasks as $S_2$, and the Process Score on open-ended tasks as $S_3$.

We implement the LLM-as-a-Judge framework\cite{llmasajudge}. The final value is derived from the arithmetic mean of scores independently assigned by \texttt{DeepSeek-V3.2}\cite{deepseekai2025deepseekv32} and \texttt{GPT-4}\cite{gpt4_technical_report_2023}. We provide a human-machine alignment study in Appendix~\ref{sec:eval_alignment}. Although human experts are typically more stringent and assign marginally lower absolute scores, they exhibit high consistency with the LLM judges in terms of both model ranking and error diagnosis. This strong correlation justifies the use of automated evaluation for large-scale assessment. Detailed prompts for both inference and evaluation are provided in Appendix~\ref{app:inference_prompt} and Appendix \ref{app:prompts}.

To quantify strict mastery, we define \textbf{$P_{obj}$} as the rate of instances achieving perfect alignment in both result and reasoning (i.e., $S_1=S_2=1.0$), and \textbf{$P_{open}$} as the rate of flawless derivation in open-ended tasks (i.e., $S_3=1.0$).

\subsection{Main Results Analysis}
\label{sec:main_results_analysis}

The main evaluation results on \textbf{OmniPhys} are presented in Table \ref{tab:main_results} and Figure \ref{fig:model_performance}. We summarize the key observations as follows:

\textbf{Dominance of Proprietary Models.} 
Proprietary systems maintain a clear advantage over open-source counterparts. Gemini-3-Pro establishes a new state-of-the-art, closely trailed by Doubao-Seed-1.6—both outperforming the flagship GPT-5.2 by over 15\%, marking a divergence in the top-tier landscape. Yet even leading models fail to saturate, with strict mastery rates ($P_{obj}$) remaining below 70\%, underscoring that OmniPhys evaluates genuine reasoning rather than rote memorization.

\textbf{Scaling Laws and Generational Gains in Open-Source Models.}
The Qwen3-VL family scales monotonically, culminating in the 235B model that surpasses the proprietary flagship GPT-5.2. Architectural superiority proves equally pivotal: Qwen3-VL-32B notably eclipses the substantially larger Qwen2.5-VL-72B. These findings confirm that algorithmic efficiency is as decisive as raw parameter scale.

\textbf{Validating Difficulty Hierarchy.}
Figure \ref{fig:model_performance} reveals a universal inverse correlation between model performance and educational stages: accuracy peaks at Junior High and degrades monotonically toward University level. This consistent drop validates the hierarchical design of OmniPhys, confirming that the benchmark captures the escalating cognitive complexity of advanced physics curricula.

\subsection{Research on Multimodal Outputs}
\label{sec:multimodal_outputs}


Current benchmarks predominantly focus on text-based tasks, overlooking the critical capacity to \textit{visualize} and \textit{construct} physical scenarios. Since drawing diagrams is a fundamental demonstration of understanding, we extend our evaluation to multimodal generation, a domain rarely explored in existing physics benchmarks.

\begin{table}[t] 
  \centering

  \resizebox{\linewidth}{!}{ 
    \begin{tabular}{l c c}
      \toprule
      \multirow{2}{*}{\textbf{Model}} & \multicolumn{2}{c}{\textbf{Evaluation Scores}} \\
      \cmidrule(lr){2-3}
      & \textbf{Human Eval.} & \textbf{Model Eval.} \\
      \midrule
      Nano Banana        & 0.23 & 0.67 \\
      Doubao-Seedream-4.0    & 0.18 & 0.59 \\
      Gemini-3-pro-Image & 0.29 & 0.73 \\
      GPT-Image-1        & 0.27 & 0.75 \\
      Qwen-Image-Edit-Plus & 0.10 & 0.51 \\
      \bottomrule
    \end{tabular}
  }
  \caption{Performance comparison in multimodal output settings.}
  \label{tab:model_drawing_scores}
\end{table}

\begin{figure}[t]
    \centering
    \includegraphics[width=\linewidth]{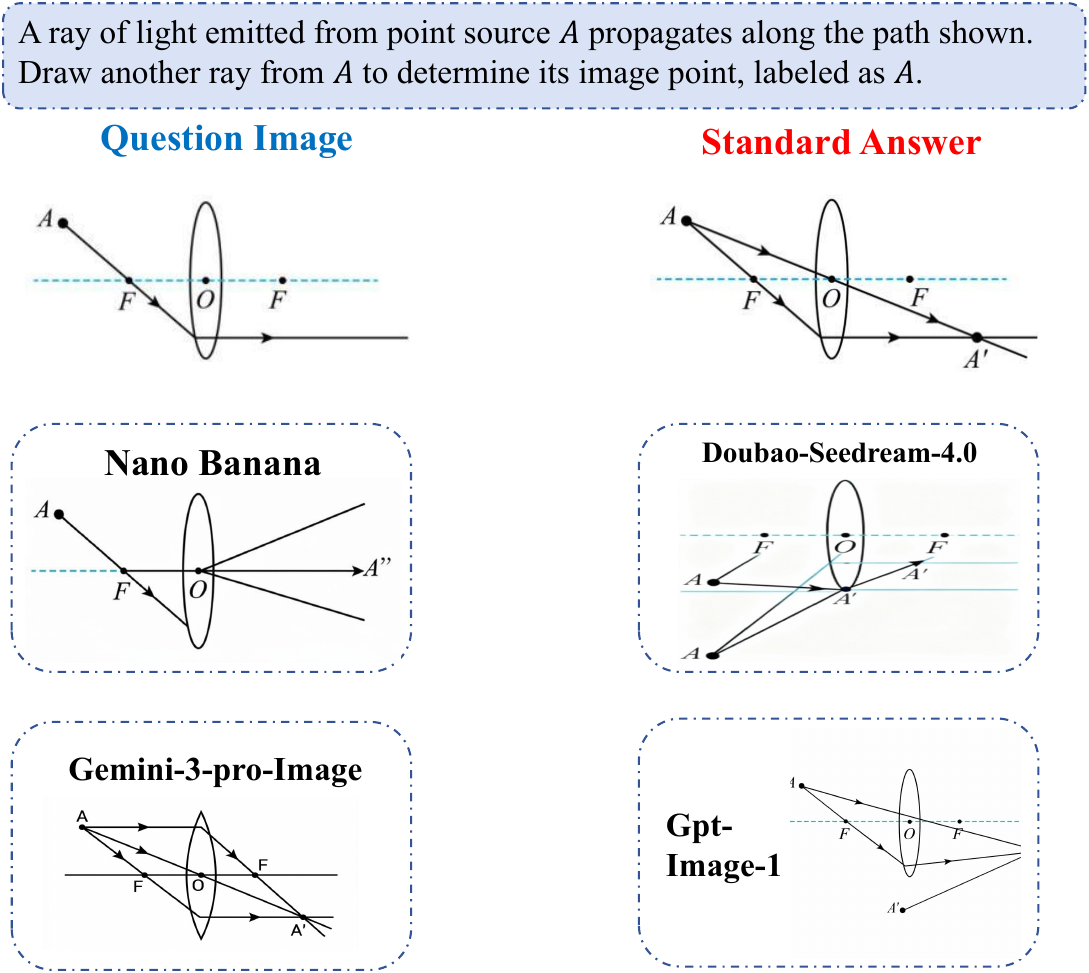}
    
    \caption{A case study of multimodal outputs in \textbf{OmniPhys} dataset.}
    \label{fig:case_study}
\end{figure}


\begin{table}[t]
    \centering
    \small
    \resizebox{\columnwidth}{!}{
    \begin{tabular}{lccc}
        \toprule
        \textbf{Model} & \textbf{Full Set} & \textbf{Test-Mini} & \textbf{Drop} \\
        \midrule
        Doubao-seed-1.6 & 76.81 & 22.45 & -70.8\% \\
        Gemini-3-pro & 80.41 & 34.66 & -56.9\% \\
        GLM-4.6v & 71.12 & 19.90 & -72.0\% \\
        GPT-5.2 & 65.60 & 21.82 & -66.7\% \\
        Qwen3-vl-plus & 68.24 & 18.76 & -72.5\% \\
        \midrule
        \textbf{Average} & \textbf{72.44} & \textbf{23.52} & \textbf{-67.8\%} \\
        \bottomrule
    \end{tabular}
    }
    \caption{Comparison of \textbf{Average Scores} (scaled to 0-100) on the full dataset versus the Test-Mini set. The significant score decline (avg. -67.8\%) confirms the \textbf{elevated difficulty} of the subset, validating its \textbf{distinct research value} as a rigorous probe for robust physical reasoning.}
    \label{tab:model_drop}
\end{table} 

To operationalize this, we define the \textbf{Physics Diagram Editing} task. Unlike standard retrieval, this requires the model to transform a multimodal input ($I_{in}, T$) into a target state ($I_{out}$) governed by strict physical laws. Our pilot experiments reveal a critical divergence: while current models exhibit high visual fidelity in general editing, they frequently violate physical constraints—failing to maintain vector directionality during coordinate transformations or preserve the topological integrity of rigid bodies. This gap underscores the necessity of our generative evaluation, exposing blind spots in physical reasoning that text-only metrics fail to capture.

\begin{table*}[t]
\centering
\small 
\setlength{\tabcolsep}{3pt}

\begin{tabular}{l cc cc cc | l cc cc} 
\toprule

\multicolumn{7}{c}{\textbf{Multimodal Large Language Models (MLLMs)}} & 
\multicolumn{5}{c}{\textbf{Large Language Models (LLMs)}} \\
\cmidrule(r){1-7} \cmidrule(l){8-12}

\multirow{2}{*}{\textbf{Model}} & \multicolumn{2}{c}{\textbf{Text+Img}} & \multicolumn{2}{c}{\textbf{Text+Caption}} & \multicolumn{2}{c}{\textbf{Text-only}} & 
\multirow{2}{*}{\textbf{Model}} & \multicolumn{2}{c}{\textbf{Text+Caption}} & \multicolumn{2}{c}{\textbf{Text-only}} \\

\cmidrule(lr){2-3} \cmidrule(lr){4-5} \cmidrule(lr){6-7} 
\cmidrule(lr){9-10} \cmidrule(lr){11-12} 

 & $P_{obj}$ & $P_{open}$ & $P_{obj}$ & $P_{open}$ & $P_{obj}$ & \multicolumn{1}{c}{$P_{open}$} & 
 & $P_{obj}$ & $P_{open}$ & $P_{obj}$ & $P_{open}$ \\
\midrule

GPT-5.2 & 2.78 & 3.02 & 4.95 & 1.08 & 5.82 & 0.86 & Deepseek-V3.2 & 5.58 & 1.08 & 5.70 & 1.73 \\
GPT-5.1 & 2.55 & 1.08 & 4.00 & 0.86 & 2.55 & 1.08 & GPT-3.5-Turbo & 1.94 & 0.63 & 1.45 & 0.22  \\
o4-mini & 7.23 & 1.96 & 6.55 & 0.86  & 6.79 & 1.73 & Qwen2.5-32B & 3.64 & 0.02 & 3.24 & 0.22 \\
Gemini-3-Pro & 25.82 & 9.50 & 16.12 & 4.10 & 15.88 & 6.26 & Qwen2.5-7B & 1.82 & 0.05 & 0.97 & 0.05 \\
Gemini-2.5-Flash & 11.88 & 2.16 & 8.12 & 1.73 & 5.70 & 1.30 & Qwen2.5-Math-7B & 1.21 & 0.22 & 1.04 & 0.02 \\
GPT-4o & 3.45 & 0.26 & 2.55 & 0.05 & 1.33 & 0.05 & InternLM-Chat-20B & 1.09 & 0.05 & 1.09 & 0.00 \\
Qwen3-VL-Plus & 3.03 & 0.43 & 5.58 & 0.22 & 7.03 & 1.30 & InternLM-Math-20B & 1.29 & 0.12 & 0.48 & 0.00 \\
GLM-4.6v & 4.00 & 0.43 & 6.42 & 0.43 & 8.12 & 1.08 & DeepSeek-R1-Distill-Qwen-7B & 2.91 & 0.12 & 4.00 & 0.05 \\
Claude-4.5-Sonnet & 3.03 & 0.65 & 2.18 & 1.30 & 3.88 & 0.86 & P1-30B-A3B & 10.67 & 1.73 & 9.82 & 2.59 \\
\midrule
\textit{\textbf{Avg.}} & \textbf{7.09} & \textbf{2.17} & \textbf{6.27} & \textbf{1.18} & \textbf{6.34} & \textbf{1.61} & \textit{\textbf{Avg.}} & \textbf{3.35} & \textbf{0.45} & \textbf{3.09} & \textbf{0.54} \\
\bottomrule
\end{tabular}
\caption{Ablation study results comparing MLLMs (left) and LLMs (right). The columns are arranged by decreasing modal information: from Text+Image to Text-only. We report Objective ($P_{obj}$) and Open-ended ($P_{open}$) Accuracy.}
\label{tab:ablation_reordered}
\end{table*}


To rigorously assess generation quality, we employed a dual-track strategy: (1) \textbf{Human Evaluation}, where three physics graduate students graded physical correctness, and (2) \textbf{Automated Evaluation}, utilizing \texttt{GPT-5.1} to score instruction adherence on a discrete $\{0, 0.5, 1\}$ scale. Detailed annotation protocols and the specific judging prompts are provided in Appendix~\ref{app:auto_eval_prompts} and Appendix~\ref{sec:appendix_interface}.

As shown in Table~\ref{tab:model_drawing_scores}, the model evaluator exhibits \textbf{excessive optimism}, consistently inflating scores ($0.51$-$0.75$) compared to human judgment ($0.10$-$0.29$). This significant divergence indicates that the current MLLM-as-a-judge paradigm lacks the robustness required for rigorous physical verification. It suggests that relying solely on automated judges is currently insufficient; specifically, the accurate discrimination of visual quality in multimodal outputs remains heavily dependent on human evaluators. Conversely, the low human ratings attest to the high difficulty and quality of our dataset, offering substantial headroom for future research and model development.

Figure~\ref{fig:case_study} presents a case study revealing that despite high visual fidelity, model-generated outputs frequently deviate from strict physical correctness. Specifically, Nano banana and Doubao-seedream-4.0 fail to correctly grasp the concept of a convex lens focus, while Gemini-3-pro-Image and gpt-Image-1 introduce erroneous, redundant line segments. These discrepancies highlight a significant capability gap in both multimodal understanding and precise generation, indicating substantial room for improvement toward human-level rigor.



\section{Ablation Study}

To enable efficient yet rigorous experimentation, we construct a \textit{Test-Mini} subset ($10\%$ of the data) using an adversarial hardness-based strategy. Specifically, samples are selected according to the consensus of five SOTA models, emphasizing high empirical failure rates ($75\%$) and reasoning complexity ($25\%$). We adopt this subset for two reasons: (1) the ablation requires evaluating each model under three input settings (Text+Img/Caption/Only), making full-set evaluation computationally expensive; and (2) the harder subset exposes modality differences more clearly, whereas easier samples in the full set tend to saturate performance and obscure cross-setting gaps. As shown in Table \ref{tab:model_drop}, this adversarial selection leads to a substantial performance drop, demonstrating strong discriminative power for ablation analysis. More details are provided in Appendix \ref{sec:appendix_mini}.

We evaluate a diverse suite of models, including deep reasoning architectures such as DeepSeek-R1-Distill-Qwen-7B and P1-30B-A3B\cite{p1-2025}, as well as variants fine-tuned on mathematics like InternLM-Math-20B\cite{cai2024internlm2} and Qwen2.5-Math-7B or physics like P1-30B-A3B. These models are assessed across three settings to quantify visual dependency: \textbf{Text+Img}, which utilizes the original multimodal input; \textbf{Text+Caption}, where diagrams are replaced by textual descriptions; and \textbf{Text-only}, which requires the model to perform blind inference based solely on the question text.

As shown in Table \ref{tab:ablation_reordered}, the adversarial \textit{Test-Mini} set imposes a significantly higher difficulty than the main benchmark, serving as a rigorous probe for deep reasoning capabilities. On average, the \textbf{Text+Img} setting yields the highest performance, with MLLMs consistently outperforming LLMs, validating the indispensable role of visual constraints in physics problems. However, we also observe a counter-intuitive phenomenon where certain models perform better in the \textbf{Text-only} setting. This suggests that for specific architectures, visual inputs or captions may be misinterpreted as distractor noise rather than helpful context, highlighting persistent challenges in cross-modal alignment that require further investigation.

\section{Conclusion}


In this paper, we present \textbf{OmniPhys}, the first unified multimodal benchmark that evaluates physics reasoning across the full educational spectrum from junior high to university, covering both understanding and generation. At its core, the DTRE protocol moves beyond surface-level answer matching by jointly scoring final results and reasoning fidelity, exposing the reasoning shortcuts that conventional metrics overlook. We further pioneer the Physics Diagram Editing task and reveal that even frontier image-generation models systematically violate physical laws, a critical blind spot invisible to text-only benchmarks. Extensive experiments across both proprietary and open-source MLLMs show that OmniPhys poses a substantial challenge, with leading models remaining below 70\% strict mastery, while ablation studies confirm the indispensable role of visual grounding. Collectively, these contributions establish OmniPhys not only as an evaluation suite, but also as a diagnostic instrument for advancing physically grounded multimodal intelligence.



\section{Limitations}

Our current work presents a foundational step with identified limitations that guide future research. 
A primary limitation lies in the linguistic and curricular scope of our data sources, which are predominantly drawn from Chinese educational materials. This monolingual setting introduces a confounding factor: weaker model performance may reflect either limited physics reasoning capability or insufficient Chinese multimodal alignment, particularly under our cross-lingual prompting protocol where instructions are in English while problems remain in Chinese. To enhance global representativeness and disentangle language effects from reasoning ability, future iterations will incorporate diverse international curricula, such as A-Level and IPhO.
Methodologies for evaluating multimodal physical outputs remain underexplored. A key challenge is the tendency of MLLM-based judges to overestimate the quality of generated content, often overlooking subtle physical inconsistencies and necessitating costly human evaluation. While we mitigate this through a structured rubric, a fully scalable solution is still lacking. We aim to establish a robust evaluation framework tailored for multimodal physics, potentially combining CV-based structural metrics with LLM semantic scoring to enable reproducible and large-scale assessment.
Finally, our current failure analysis is limited in depth, offering only qualitative observations of representative error patterns. We plan to conduct a rigorous error attribution study at scale, systematically distinguishing visual perception failures from logical reasoning errors, thereby providing more granular insights into MLLM mechanisms in physics reasoning.

\section*{Acknowledgments}
This study was funded by Guangxi Science and Technology Program (2025AB25069309), the Open Research Fund of Key Laboratory of Advanced Theory and Application in Statistics and Data Science (East China Normal University).

\bibliography{custom}

\appendix

\section{Prompt Usage}

All prompts presented in this section are translated into English for readability. In our actual experiments, prompts are administered in Chinese to align with the language of the OmniPhys benchmark, ensuring consistent cross-modal understanding for both inference and evaluation.

\subsection{Visual Dependency Annotation}
\label{app:visual_dependency_prompt}

\begin{tcolorbox}[colback=blue!5!white, colframe=blue!40!black, title= Prompt for Visual Dependency Annotation, before upper=\setstretch{1.2},enhanced jigsaw,
    breakable, ]

You are an analyst of multimodal physics problem datasets. You are given access only to the problem text, without seeing the accompanying image. Based on the textual description, infer the importance of the image for solving the problem.

Image importance levels are defined as follows:

\textbf{Level 1 (Text-Only Solvable)}:
The image is purely decorative or illustrative. All information required to solve the problem is fully described in the text, and the image is not necessary.

\textbf{Level 2 (Text-Descriptive)}:
The image contains information, but the text fully restates or describes all visual content. The image serves only as an aid for understanding and is not strictly required for reasoning.

\textbf{Level 3 (Image-Essential)}:
The text refers to the image (e.g., "as shown in the figure"), and critical information such as numerical values, geometric relationships, or measurement readings is available only in the image. The image is essential for solving the problem.

Please output only a single integer (1, 2, or 3) corresponding to the image importance level.

Problem text:
\{problem\_text\}
\end{tcolorbox}

\subsection{Model Inference Prompt}
\label{app:inference_prompt}

\begin{tcolorbox}[colback=blue!5!white, colframe=blue!40!black, title= Prompt for Model Inference, before upper=\setstretch{1.2},enhanced jigsaw,
    breakable, ]

I will present you with a physics problem. Please read and solve it. Adhere strictly to the following output format:

'''
Reasoning Process: [Solve step-by-step logically. Limit each step to no more than 30 words, focusing only on core deductions without redundant explanations.]

Answer: [Output your final answer. Do not add extra content.]
'''

[Input Image]

\{problem\_text\}
\end{tcolorbox}

\subsection{Automated Evaluation Prompts}
\label{app:prompts}

We employ a dynamic prompting strategy based on the question type. The evaluator (LLM-as-a-Judge) receives specific instructions for objective and open-ended tasks respectively.

\begin{tcolorbox}[colback=blue!5!white, colframe=blue!40!black, title= Prompt for Objective Tasks (Dual-Track Evaluation), before upper=\setstretch{1.1}, enhanced jigsaw, breakable]

You are a rigorous grader. Please conduct a dual-track evaluation for this objective task: assess both the correctness of the final result and the validity of the reasoning process.

\textbf{[Question Data]}
[Question]: \{question\}
[Standard Answer]: \{std\_answer\}
[Explanation]: \{std\_explanation\}

\textbf{[Student Response]}
\{model\_answer\}

\textbf{[Evaluation Tasks]}
Please complete the following two parts and merge them into a JSON output:

1. \textbf{Process Analysis (process\_eval)}:
   - Decompose the standard solution into $m$ key steps.
   - Count how many steps ($n$) the student correctly completed.
   - Process Score = $n / m$.
   - If the student provides the correct answer without reasoning steps, treat it as a "reasoning shortcut" (assign score based on context, typically penalized or full if trivial).

2. \textbf{Result Analysis (result\_eval)}:
   - Ignore the process and strictly check if the final answer matches the standard.
   - For Single Choice / True-False: 1.0 for match, 0.0 for mismatch.
   - For Multi-Select / Fill-in-the-Blank: Score = (Count of Correct Slots) / (Total Required Slots).
   - Ignore format variations (e.g., "A" vs. "A.").

\textbf{[Output Requirement]}

Output ONLY in JSON format without Markdown tags:
\{
    "process\_eval": \{
        "reason": "Brief analysis (Total m steps, Correct n steps)",
        "total\_steps": m,
        "correct\_steps": n,
        "process\_score": 0.0 to 1.0
    \},
    "result\_eval": \{
        "reason": "Brief justification for the result",
        "score": 0.0 to 1.0
    \}
\}
\end{tcolorbox}

\vspace{1em}

\begin{tcolorbox}[colback=green!5!white, colframe=green!40!black, title= Prompt for Open-Ended Tasks (Process-Only Evaluation), before upper=\setstretch{1.1}, enhanced jigsaw, breakable]

You are an expert physics evaluator. Your task is to assess the student's reasoning logic step-by-step.

\textbf{[Question Data]}

[Question]: \{question\}
[Standard Answer]: \{std\_answer\}
[Explanation]: \{std\_explanation\}

\textbf{[Student Response]}

\{model\_answer\}

\textbf{[Scoring Criteria]}
1. Decompose the complete resolution process into $m$ \textbf{Key Reasoning Steps}.
2. Determine how many of these steps ($n$) are correctly included in the student's response.
3. The final Process Score is $n / m$.

\textbf{[Output Requirement]}

Output ONLY in JSON format without Markdown tags:
\{
    "reason": "Step-by-step analysis of the derivation",
    "total\_steps": m (integer),
    "correct\_steps": n (integer),
    "process\_score": Calculated result of n/m (0.0 to 1.0)
\}
\end{tcolorbox}

\subsection{Automated Evaluation Prompts for Multimodal Generation}
\label{app:auto_eval_prompts}

We utilize a Vision-Language Model (e.g., GPT-5.1 or GPT-4o) as the evaluator to assess the quality of generated physics diagrams. The judge receives the problem text, the context image (if available), and the generated diagram, then assigns a discrete score based on physical fidelity.

\begin{tcolorbox}[colback=blue!5!white, colframe=blue!40!black, title= Prompt for Diagram Editing Evaluation, before upper=\setstretch{1.1}, enhanced jigsaw, breakable]

You are a professional and rigorous physics instructor. Your task is to grade physics diagrams drawn by students based on specific problem requirements.

\textbf{[Input Data]}

1. \textbf{Problem Description}: Describes the physical scenario or process to be drawn.
2. \textbf{Context Image} (Optional): Provides background information (if any).
3. \textbf{Student Answer Image}: The diagram generated by the student.

\textbf{[Evaluation Task]}

Based on physical principles, comprehensively evaluate whether the student's image accurately and completely fulfills the problem requirements. The diagrams may involve mechanics analysis, motion trajectories, optical paths, circuit connections, or electromagnetic field distributions.

\textbf{[Scoring Criteria]}
\begin{itemize}
    \item \textbf{1.0 (Fully Correct)}: The image perfectly reflects the problem requirements. All key physical elements (e.g., vector directions, points of application, trajectory shapes, light paths, circuit connections, physical labels) are accurate and adhere to physical laws.
    \item \textbf{0.5 (Partially Correct)}: The image captures the core physical concept but contains defects in details. Examples: Main structure is correct but minor labels are missing; key vectors are roughly correct in direction but deviate significantly in angle or proportion; general trend is correct but local errors exist; or unnecessary misleading lines are included.
    \item \textbf{0.0 (Incorrect)}: The image contains fundamental physical errors or omits the most critical information, failing to reflect the problem requirements. Examples: Depicting the wrong physical phenomenon; missing core elements required by the stem; or generating an image completely irrelevant to the problem.
\end{itemize}

\textbf{[Output Requirement]}
Please output strictly in JSON format (no Markdown tags):
\{
    "reasoning": "Concise justification pointing out specific merits or demerits",
    "score": 1.0 or 0.5 or 0.0
\}

\textbf{[Input Sequence]}
[Problem Description]: \{question\_text\}

[Context Image]: (Input Image Here)

[Student Answer Image]: (Generated Image Here)
\end{tcolorbox}

\begin{figure*}[hbt!]
    \centering
    \includegraphics[width=0.95\textwidth]{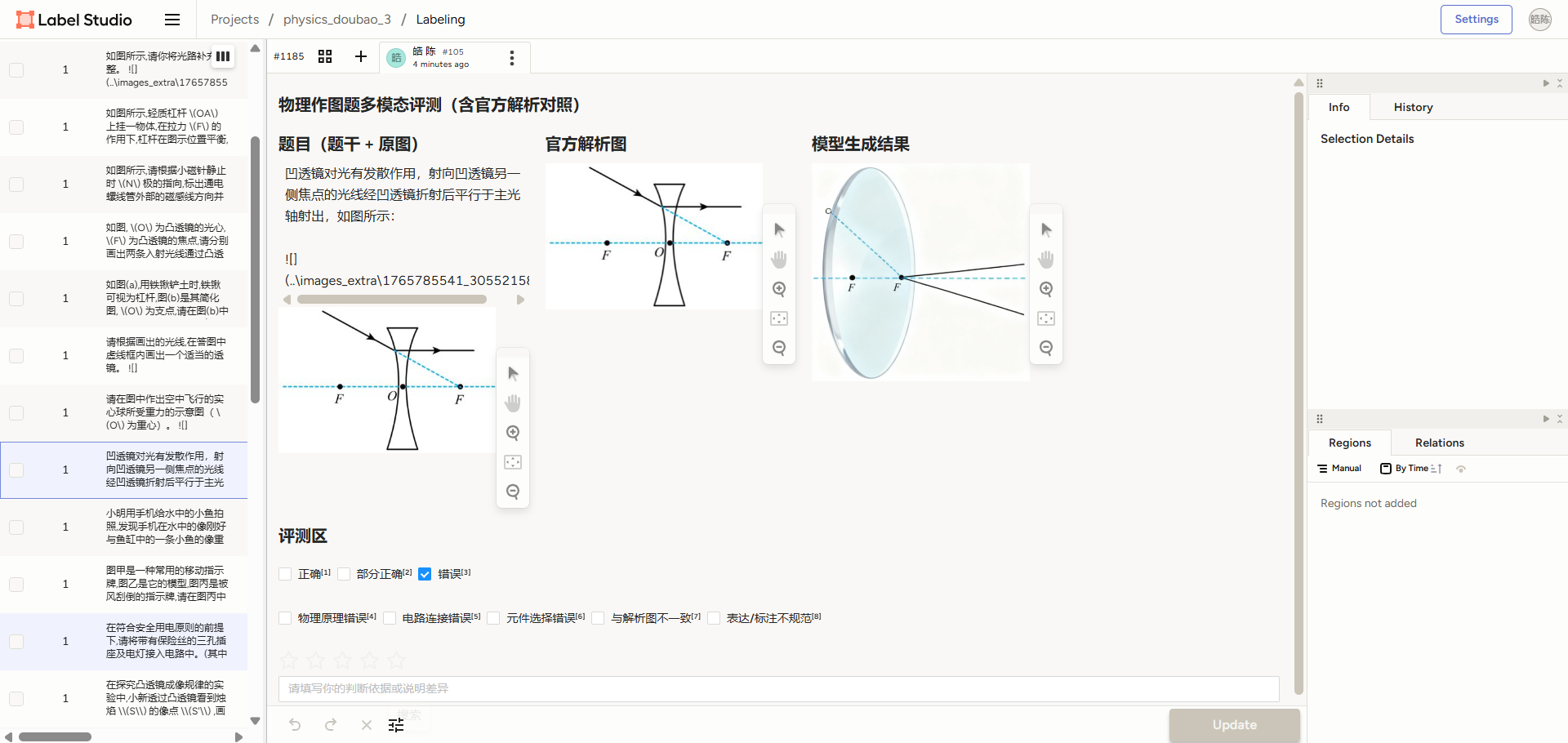}
    \caption{Screenshot of the Label Studio interface for human evaluation of multimodal outputs.}
    \label{fig:label_studio}
\end{figure*}

\section{Annotation Interface Details}
\label{sec:appendix_interface}

To ensure the quality and consistency of our evaluation, we utilized the Label Studio platform for manual annotation. Figure~\ref{fig:label_studio} illustrates the annotation interface used by human evaluators to assess the model's multimodal outputs.

The interface was carefully designed to present all relevant information required for reliable judgment within a single view. Specifically, it simultaneously displays (1) the original problem statement, (2) the original input image, (3) the ground-truth reference image provided by the dataset, and (4) the image generated by the evaluated model. This design allows annotators to directly compare the model output with the reference solution in the context of the original task, thereby making the scoring process both efficient and less prone to omission or misinterpretation.

For each sample, annotators were asked to assign a quality score based on the correctness, completeness, and visual faithfulness of the generated image with respect to the reference answer. Since all relevant inputs and outputs are visible in a unified interface, the evaluation process is straightforward to operate and reduces unnecessary cognitive load on the annotators.

The annotation was conducted by three graduate students specializing in physics, all of whom had prior experience with problem solving and diagram interpretation in the target domain. To improve reliability, each sample was independently annotated by all three evaluators, and the final human evaluation score reported in our experiments is computed as the average of the three scores. This aggregation strategy helps mitigate individual bias and increases the robustness of the evaluation results. The annotators were compensated with standard research assistant stipends in accordance with institutional guidelines.




\section{Test-Mini Construction and Ablation Details}
\label{sec:appendix_mini}

In this section, we provide a detailed elaboration on the construction process of the \textit{Test-Mini} set and the specific settings used for the ablation study.

\subsection{Hardness-Based Selection Methodology}
To ensure the \textit{Test-Mini} set effectively probes the upper limits of multimodal reasoning, we implemented a multi-dimensional hardness scoring mechanism. For each candidate question $x_i$ in the full dataset (Without multimodal outputs, $N=12,885$), we computed a hardness score $H(x_i)$ based on two factors: empirical model failure and reasoning complexity.

\begin{table}[ht]
    \centering
    \small
    \resizebox{\columnwidth}{!}{
    \begin{tabular}{lrr}
        \toprule
        \textbf{Metric} & \textbf{Full Set} & \textbf{Test-Mini} \\
        \midrule
        Sample Size & 12,885 & 1,288 \\
        Avg. Reasoning Length (chars) & 563 & 979 \\
        Visual Necessity Score & 8.98 & 8.99 \\
        All-Model Failure Rate & 1.6\% & 15.5\% \\
        \textbf{Avg. Model Accuracy} & \textbf{72.44\%} & \textbf{23.52\%} \\
        \bottomrule
    \end{tabular}
    }
    \caption{Comparison of key statistics between the full dataset and the selected Test-Mini subset. The subset demonstrates significantly higher complexity and lower model solvability.}
    \label{tab:dataset_stats}
\end{table}

The score is defined as:
\begin{equation}
    H(x_i) = w_1 \cdot \mathcal{F}(x_i) + w_2 \cdot \mathcal{C}(x_i)
\end{equation}
where:
\begin{itemize}
    \item $\mathcal{F}(x_i)$ represents the \textbf{Empirical Failure Rate}. It is calculated as $\mathcal{F}(x_i) = 1 - \frac{1}{|M|} \sum_{m \in M} S(m, x_i)$, where $M$ is the set of five baseline models and $S(m, x_i) \in [0,1]$ is the normalized score of model $m$ on question $x_i$.
    \item $\mathcal{C}(x_i)$ represents the \textbf{Reasoning Complexity}, quantified by the percentile rank of the character length of the ground truth explanation (Chain-of-Thought).
    \item We set the weights to $w_1 = 0.75$ and $w_2 = 0.25$, prioritizing empirical difficulty while accounting for logical depth.
\end{itemize}

The baseline models set $M$ includes five state-of-the-art closed-source models: \textit{Doubao-seed-1.6}, \textit{Gemini-3-pro-preview}, \textit{GLM-4.6v}, \textit{GPT-5.2}, and \textit{Qwen3-vl-plus}. We selected the top 10\% of samples ranked by $H(x_i)$ to form the \textit{Test-Mini} set.

\subsection{Statistics and Performance Gap}
The selection process resulted in a subset with significantly higher difficulty. As shown in Table \ref{tab:dataset_stats}, the \textit{Test-Mini} set exhibits a sharp increase in the "All-Model Failure Rate" (questions where no model answered correctly) from 1.6\% to 15.5\%, and a substantial increase in the average reasoning chain length.

Table \ref{tab:model_drop} details the performance drop for each baseline model. The consistent degradation across all models (ranging from 56.9\% to 72.5\%) confirms that the difficulty of the \textit{Test-Mini} set is not biased towards a specific architecture but stems from the inherent complexity of the physics problems.

\section{Representative Source Materials}
\label{sec:appendix_samples}
\begin{figure*}
    \centering
    \includegraphics[width=0.95\linewidth]{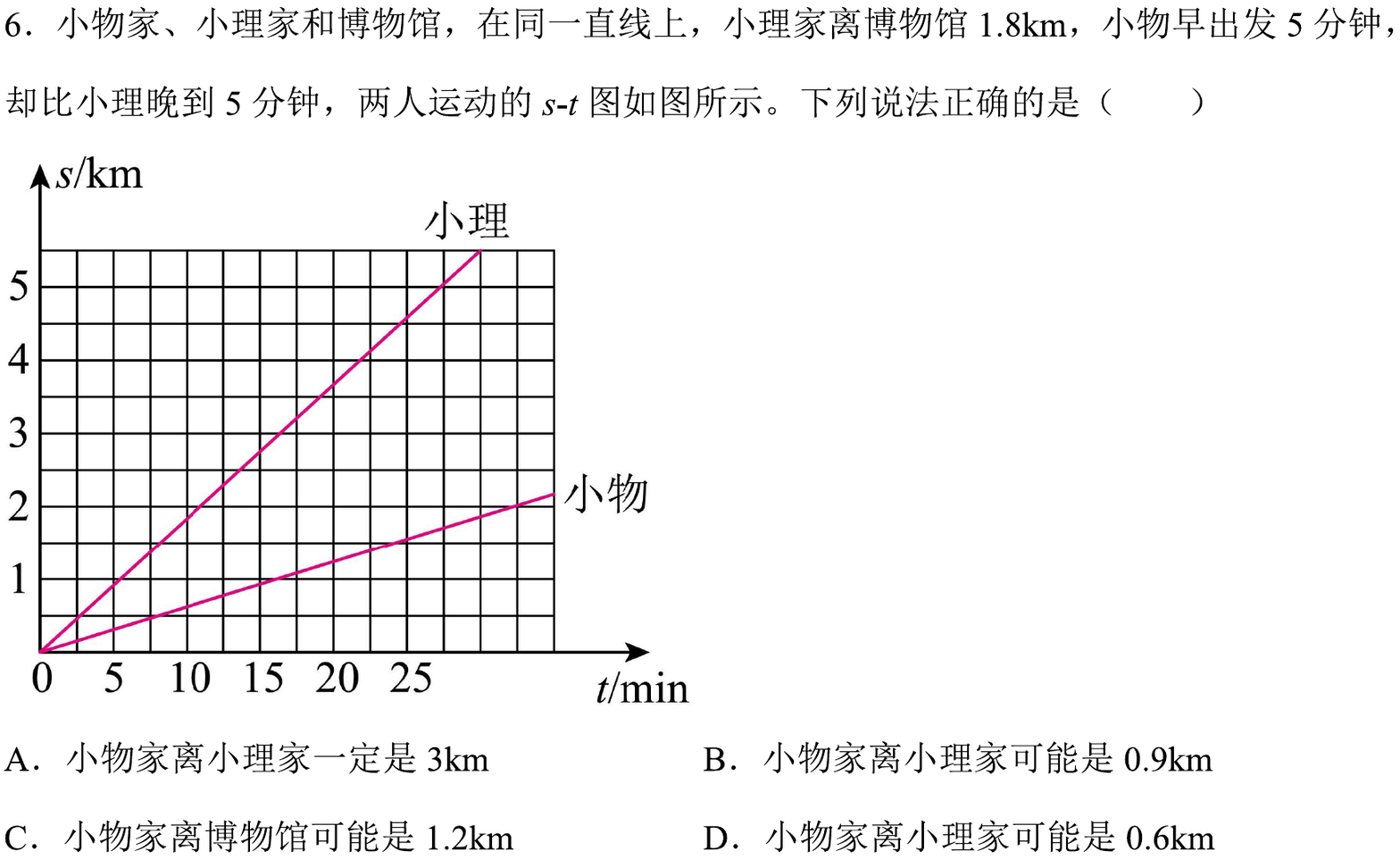}
    \caption{A middle school, Mechanics example of Original Physics Problems.}
    \label{fig:example_1}
\end{figure*}

\begin{figure*}
    \centering
    \includegraphics[width=0.95\linewidth]{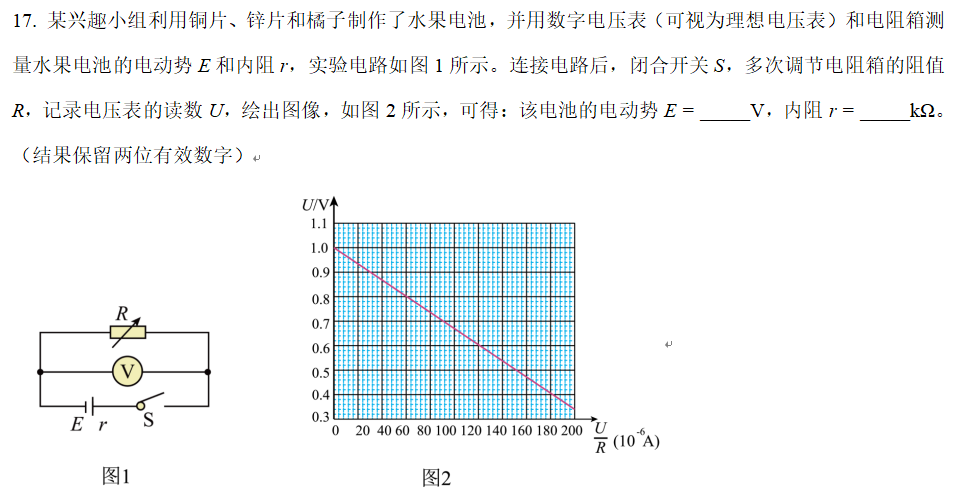}
    \caption{A high school, Electromagnetism example of Original Physics Problems.}
    \label{fig:example_2}
\end{figure*}

\begin{figure*}
    \centering
    \includegraphics[width=0.95\linewidth]{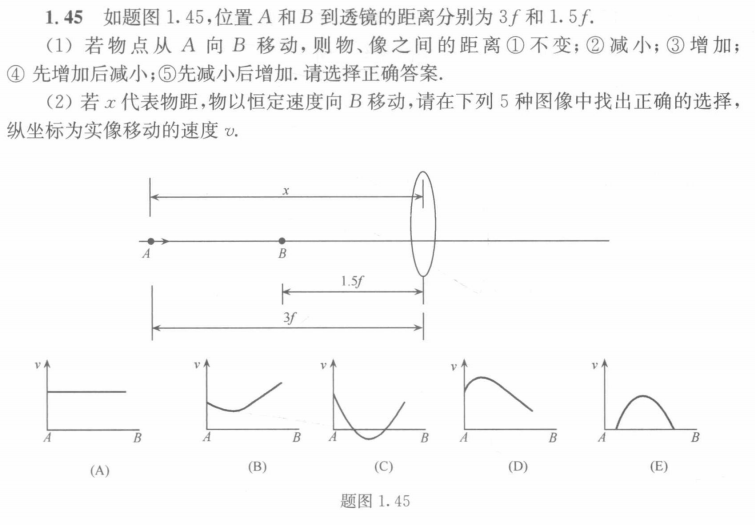}
    \caption{A University, Optics example of Original Physics Problems.}
    \label{fig:example_3}
\end{figure*}

To demonstrate the authenticity and complexity of our data, we provide a selection of original screenshots from the standardized Chinese physics examination papers and authoritative textbooks used to curate the \textbf{OmniPhys} benchmark. Figure \ref{fig:example_1}, Figure \ref{fig:example_2} and Figure \ref{fig:example_3} show 3 examples of original physics problems from Chinese real-world exams.

\section{Human-Machine Alignment and Quality Control in Data Filtering Process}
\label{sec:alignment}

To address potential biases in the automated components of our pipeline—specifically Visual Dependency Filtering and Difficulty Screening—we conducted a blind correlation study involving two physics experts (Ph.D. candidates). We randomly sampled 300 instances for manual review.

\begin{table*}[t]
    \centering
    \small
    \renewcommand{\arraystretch}{1.1}
    \begin{tabular}{l c c c}
        \toprule
        \textbf{Filtering Task} & \textbf{Metric} & \textbf{Human-Human} & \textbf{Human-Machine} \\
        \midrule
        \multirow{2}{*}{Visual Dependency} & Agreement (\%) & 92.6\% & 89.3\% \\
                                           & Cohen's $\kappa$ & 0.86 & 0.82 \\
        \midrule
        \multirow{2}{*}{Difficulty Screening} & Agreement (\%) & 94.1\% & 91.5\% \\
                                           & Cohen's $\kappa$ & 0.88 & 0.84 \\
        \bottomrule
    \end{tabular}
    \caption{\textbf{Alignment Statistics for Data Refinement.} The "Human-Machine" column represents the consistency between our automated pipeline (using MLLMs and vLLM) and expert consensus. Scores above 0.80 signify strong reliability in our data quality control.}
    \label{tab:alignment_extended}
\end{table*}

\begin{table*}[h]
\centering
\small

\begin{tabular}{l cccc}
\toprule
\textbf{Metric} & \textbf{Pearson $r$} & \textbf{Spearman $\rho$} & \textbf{Agreement (\%)} & \textbf{$\Delta$ Score} \\
\midrule
$S_1$ (Answer Accuracy) & 0.89 & 0.87 & 92.4\% & +1.2\% \\
$S_2$ (Obj. Process)   & 0.86 & 0.84 & 89.7\% & +2.5\% \\
$S_3$ (Subj. Process)  & 0.82 & 0.81 & 85.5\% & +4.2\% \\
\midrule
$P_{obj}$ (Mastery)    & 0.92 & 0.91 & 93.1\% & +0.8\% \\
$P_{open}$ (Mastery)   & 0.88 & 0.86 & 91.2\% & +1.5\% \\
\bottomrule
\end{tabular}
\caption{\textbf{Human-Machine Alignment Results.} Comparison between expert consensus and the LLM-as-a-Judge framework across primary metrics. $\Delta$ Score represents the mean difference (Machine $-$ Human).}
\label{tab:alignment_results}
\end{table*}

The high alignment scores in Table \ref{tab:alignment_extended} demonstrate that our adversarial filtering strategy effectively targets non-trivial problems while maintaining the pedagogical integrity of the physics domains.

\section{Human-Machine Alignment on Evaluation}
\label{sec:eval_alignment}

To validate the reliability of the LLM-as-a-Judge framework (comprising \texttt{DeepSeek-V3} and \texttt{GPT-4}) used in our main experiments, we conducted a systematic alignment study. We randomly sampled 300 model-generated responses across five physics domains and three task types. Three physics experts (graduate students) independently scored these responses using the same rubric as the LLM judges.

Table \ref{tab:alignment_results} presents the correlation and agreement metrics. Our analysis reveals three key findings:
\begin{itemize}
    \item \textbf{Strong Ranking Consistency}: The Pearson correlation ($r$) for $S_1$ (Accuracy) and $S_2$ (Objective Process) exceeds 0.85, indicating that the LLM judges and human experts are highly consistent in ranking model performance across different physics domains.
    \item \textbf{Expert Stringency}: We observe that human experts are typically more stringent, resulting in absolute scores marginally lower than the LLM consensus (with a mean difference $\Delta$ of 1.2\% to 4.2\%). This is primarily due to experts' higher sensitivity to subtle conceptual inaccuracies in complex reasoning chains.
    \item \textbf{Robust Mastery Detection}: The agreement on strict correctness rates ($P_{obj}$ and $P_{open}$) reaches 91.2\%, confirming that the "perfect performance" threshold used in Table \ref{tab:main_results} provides a reliable signal for identifying model mastery.
\end{itemize}

Despite the minor bias in absolute values, the strong correlation across all dimensions justifies the scalability of our automated evaluation protocol for the large-scale assessment in OmniPhys.

\section{AI Assistant Usage Disclosure}
\label{sec:ai_disclosure}
We used AI assistants (Gemini/ChatGPT) to assist with writing script code for data processing and for polishing the language of the manuscript to improve readability. All scientific claims, experimental designs, and final text were manually verified and revised by the authors.

\end{document}